\pdfoutput=1

\documentclass[11pt]{article}

\usepackage{ACL2023}

\usepackage{times}
\usepackage{latexsym}

\usepackage[T1]{fontenc}

\usepackage[utf8]{inputenc}

\usepackage{microtype}

\usepackage{inconsolata}

\title{Did the Models Understand Documents? Benchmarking Models for Language Understanding in Document-Level Relation Extraction}

\author{Haotian Chen, Bingsheng Chen \and Xiangdong Zhou \\
        School of Computer Science, Fudan University \\ Shanghai Key Laboratory of Data Science \\ \texttt{htchen18@fudan.edu.cn, chenbs21@m.fudan.edu.cn, xdzhou@fudan.edu.cn} }

\usepackage{bibentry}
\usepackage{epsfig}
\usepackage{multicol}
\usepackage{multirow}
\usepackage{booktabs}
\usepackage{amsmath}
\usepackage{amssymb}
\usepackage{makecell}
\usepackage{color}
\usepackage{algorithm}
\usepackage{algorithmic}
\usepackage{graphics}
\usepackage{wrapfig}
\usepackage{bm}

\graphicspath{{figure/}}

\begin{document}

\maketitle

\begin{abstract}
Document-level relation extraction (DocRE) attracts more research interest recently. While models achieve consistent performance gains in DocRE, their underlying decision rules are still understudied: Do they make the right predictions according to rationales? In this paper, we take the first step toward answering this question and then introduce a new perspective on comprehensively evaluating a model.
Specifically, we first conduct annotations to provide the rationales considered by humans in DocRE. Then, we conduct investigations and reveal the fact that: In contrast to humans, the representative state-of-the-art (SOTA) models in DocRE exhibit different decision rules. Through our proposed RE-specific attacks, we next demonstrate that the significant discrepancy in decision rules between models and humans severely damages the robustness of models and renders them inapplicable to real-world RE scenarios. After that, we introduce mean average precision (MAP) to evaluate the understanding and reasoning capabilities of models. According to the extensive experimental results, we finally appeal to future work to consider evaluating both performance and the understanding ability of models for the development of their applications. We make our annotations and code publicly available\footnote{\url{https://github.com/Hytn/DocRED-HWE}}.
\end{abstract}


\section{Introduction}
Relation extraction (RE), aiming to extract relations between entities from texts, plays an important role in constructing a large-scale knowledge graph~\cite{riedel2010modeling,hendrickx2010semeval}. Most previous work extract relations from a single sentence~\cite{zelenko2002kernel,wei2020novel,shang2022onerel}, while recent studies adopt multiple sentences as a whole to harvest more relations including inter-sentence relations~\cite{yao2019docred}, i.e., document-level relation extraction (DocRE). DocRE is more challenging because models are required to synthesize all information of a given document and then predict relations by reasoning and language understanding~\cite{yao2019docred,nan2020reasoning,zeng2020double}.  

\begin{figure}
  \centering
  \includegraphics[width=1.0\columnwidth]{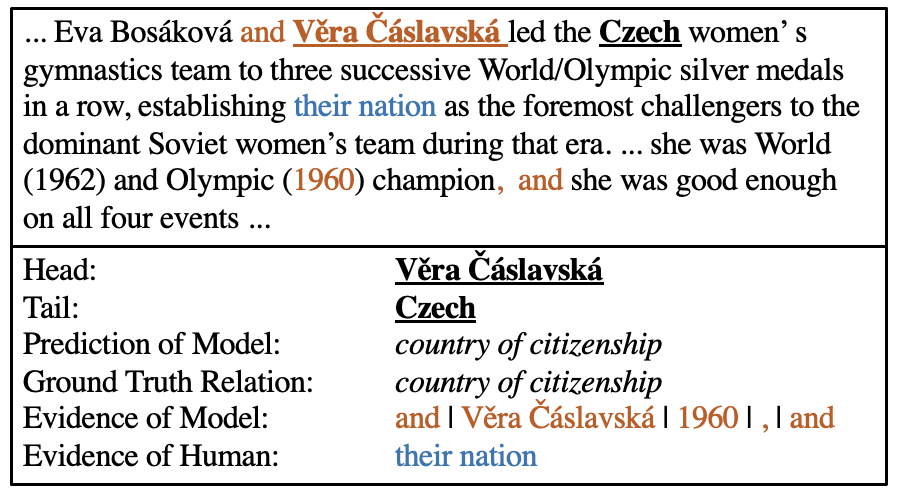}
  \caption{An example from DocRED.}
  \label{fig:IG_viz-fig}
\end{figure}

Previous work obtains consistent performance gains on DocRED~\cite{yao2019docred}, the proposal of which has benefited the rapid development of DocRE~\cite{huang2022does}. However, the extent to which their proposed methods possess language understanding and reasoning capabilities is still understudied. A common evaluation method is to measure average error across a test set, which neglects the situations where models can make right predictions according to wrong features. As shown in Figure~\ref{fig:IG_viz-fig}, the model accurately predicts the relation between \textit{Věra Čáslavská} and \textit{Czech} as humans do. However, the evidence words considered by models are incomprehensible to humans. 
Similar situations, where models improve their performance by recognizing the spurious patterns, are identified by parts of the AI community, including \emph{annotation artifacts} in natural language inference (NLI)~\cite{poliak2018hypothesis,gururangan2018annotation,glockner2018breaking} and \emph{shallow template matches} in named entity recognition (NER)~\cite{fu2020rethinking}. These learned spurious patterns can severely damage their robustness and generalization abilities in the corresponding tasks~\cite{geirhos2020shortcut}. To the best of our knowledge, this is the first work to diagnose the decision rules of models in DocRE.
In this paper, we analyze and characterize the understanding ability of SOTA models in DocRE, expose the bottleneck of the models, and then introduce a new evaluation metric to select trustworthy and robust models from those well-performed ones. Our contributions are summarized as follows:

(1) We conduct careful and exhausting annotations on DocRED to propose $\text{DocRED}_\text{HWE}$, where HWE denotes human-annotated word-level evidence. The evidence words (decision rule) of humans are annotated in the dataset. 

(2) We adopt a feature attribution method to observe the most crucial words considered by models in their reasoning processes. We reveal that the SOTA models spuriously correlate the irrelevant (non-causal) information (e.g., entity names, some fixed positions in any given documents, and irrelevant words) with their final predictions, forming their own unexplainable decision rules.

(3) We demonstrate that the decision rules of the SOTA models in DocRE are not reliable. We delicately design six kinds of RE-specific attacks to expose their bottleneck: Although they succeed in achieving improved performance on the held-out test set, they can strikingly fail under our designed attacks. Both the severe lack of understanding ability and the learned spurious correlations exacerbate the vulnerability of the models. 

(4) Inspired by evaluation metrics in recommender systems, we evaluate the understanding and reasoning capability of models by our introduced mean average precision (MAP). MAP enables us to distinguish between the spurious-correlation-caused and the understanding-ability-caused improvements in the performance of models. We observe that a model with a higher MAP will achieve stronger robustness and generalization ability.
\section{Related Work}

\paragraph{Document-level Relation Extraction.}
Prevalent effective methods on document-level RE can be divided into two categories: graph-based methods and transformer-based methods~\cite{huang2022does}. Both of them are based on deep neural networks (DNNs). Graph-based methods explore the structure information in context to construct various graphs and then model the process of multi-hop reasoning through the paths in graphs. According to the classification mentioned in previous work~\cite{huang2022does}, the SOTA graph-based method is DocuNet~\cite{zhang2021document}, which constructs an entity-level relation graph, and then leverages a U-shaped network over the graph to capture global interdependency. 
Transformer-based methods perform reasoning by implicitly recognizing the long-distance token dependencies via transformers. One of the most representative transformer-based methods is ATLOP~\cite{zhou2020documentlevel}, which enhances the embeddings of entity pairs by relevant context and introduces a learnable threshold for multi-label classification. The techniques proposed by ATLOP are widely adopted by subsequent transformer-based work~\cite{xie2022eider,tan2022documentlevela,xiao2022sais}, including adaptive thresholding (AT) and localized context pooling (LOP). 

\paragraph{Analyzing Decision Rules of DNNs.}
With the tremendous success and growing societal impact of DNNs, understanding and interpreting the behavior of DNNs has become an urgent necessity. In terms of NLP, While DNNs are reported as having achieved human-level performance in many tasks, including QA~\cite{chen2019codah}, sentence-level RE~\cite{wang2020tplinker}, and NLI~\cite{devlin2018bert}, their decision rules found by feature attribution (FA) methods are different from that of humans in many cases. For example, in argument detection, the widely adopted language model BERT succeeds in finding the most correct arguments only by detecting the presence of ``not''~\cite{niven2019probing}. In VQA, dropping all words except ``color'' in each question is enough for a DNN to achieve $50\%$ of its final accuracy~\cite{mudrakarta2018dida}. In NLI, DNNs can make the right predictions without access to the context~\cite{poliak2018hypothesis}.
It is demonstrated in these tasks that decision rules of models should approach that of humans. Otherwise, the difference will lead to a severe lack of robustness and generalization ability~\cite{agrawal2016analyzing,belinkov2018synthetic,fu2020rethinking}. It remains understudied whether the same conclusion is established in DocRE. To the best of our knowledge, this is the first work comprehensively analyzing the decision rules of both models and humans in DocRE.
\section{Data Collection}
Our ultimate goal is to provide all of the evidence words (decision rules) that humans rely on during the reasoning process in DocRE. Since it is not feasible for annotators to label relations and evidence from scratch in DocRE~\cite{yao2019docred,huang2022does}, we select DocRED to further annotate our fine-grained decision rule (word-level evidence). Our proposed dataset is named $\text{DocRED}_\text{HWE}$, where $\text{HWE}$ denotes human-annotated word-level evidence. In the following two sections, we first elaborate on the underlying reasons why we conduct word-level evidence annotation and why on DocRED, and then introduce the details of our annotation. 

\subsection{Motivations}
\label{motivations}
\paragraph{Motivation for Human Annotation.}
Current human annotations on DocRED are still insufficient to support our research: the evidence for each relational fact is sentence-level instead of word-level. If we base our study on the coarse-grained decision rules (sentence-level evidence) to analyze the reasoning behaviors of humans and models, the results will be misleading. For example, as shown in Figure~\ref{fig:IG_viz-fig}, the sentence-level evidence of models and humans overlaps with each other (\textit{and Věra Čáslavská} and \textit{their nation} come from the same sentence), while their word-level evidence is totally different. Therefore, annotation of word-level evidence is of the essence. we conduct careful and exhausting word-level evidence annotation on DocRED and propose $\text{DocRED}_\text{HWE}$. Our proposed dataset significantly benefits more comprehensive analyses of DocRE, which will be discussed in Section~\ref{experiments}.
\paragraph{Motivation for Selecting DocRED.}
While there are a few candidate datasets in DocRE, only one of them named DocRED~\cite{yao2019docred} satisfies the urgent need of studying the understanding and reasoning capabilities of general-purpose models in real-world DocRE. Specifically, \citet{quirk2017distant} and \citet{peng2017crosssentencea} leverage distant supervision to construct two datasets without human annotation, which hurts the reliability of the evaluation. \citet{li2016biocreative} and \citet{wu2019renet} proposed two human-annotated document-level RE datasets named CDR and GDA, respectively. Both of them serve specific domains and approaches (biomedical research) and contain merely one to two kinds of domain-specific relations. Different from other datasets in DocRE, the proposal of DocRED has significantly promoted the rapid development of the task in the past two years~\cite{huang2022does}. The large-scale human-annotated dataset is constructed from Wikipedia and Wikidata, which serves general-purpose and real-world DocRE applications~\cite{yao2019docred}. Among various improved versions of DocRED~\cite{huang2022does,tan2022revisiting}, we select the original version with annotation noise because it presents one of the most general circumstances faced by RE practitioners: having limited access to entirely accurate human-annotated data due to the extremely large annotation burden and difficulty. For example, human-annotated DocRED and TACRED~\cite{zhang2017positionawarea} are discovered to have labeling noise. As to distantly supervised datasets NYT~\cite{mintz2009distant} and DocRED-distant, the amount of noise becomes larger.

\subsection{Human Annotation Generation}
\paragraph{Challenges and Solutions.}
We randomly sample 718 documents from the validation set of DocRED. Annotators are required to annotate all the words they rely on when reasoning the target relations. Note that we annotate the pronouns that can be another kind of mentions for each entity, which are crucial for logical reasoning but neglected in DocRED. 
Our annotation faces two main challenges. \textit{The first challenge} comes from the annotation artifacts in the original dataset: Annotators can use prior knowledge to label the relations through entity names, without observing the context. For example, given a document with a cross-sentence entity pair ``Obama'' and ``the US'', annotators tend to label ``president of'' despite the lack of rationales. The issue is naturally solved by annotating the fine-grained word-level evidence. Consequently, despite the intensive workload, we annotate the words in reasoning paths for each relation.
\textit{The second challenge} lies in multiple reasoning paths for a single relation: Annotators are required to annotate the words in all reasoning paths. While annotators succeed in reasoning a certain relation through the corresponding evidence words, those words in other reasoning paths can often be neglected. To solve the issue, we adopt multiple (rolling) annotations for each document and propose the checking rule: Given a document and the previously annotated relation with its evidence words masked, the annotator will not be able to reason the relation. If the rule is violated, new evidence words will be annotated. The update will be checked by the next annotator until no update occurs. All of the annotated evidence words are verified at least two times.

\paragraph{Quality of Annotation.}
To ensure the quality of the dataset, we provide principle guidelines and training to the annotators. We examine the annotators if they understand the principle. Meanwhile, we regularly inspect the quality of annotations produced by each annotator. Our inspection exerts a positive effect on the quality. On one hand, we filter out 18 out of 718 documents that present low annotation accuracy. Through the rolling annotation strategy, annotators also inspect the annotations from each other.
On the other hand, annotators correct three kinds of annotation errors in the original DocRED: 1) relation type error where annotators wrongly annotate a relation type between an entity pair, 2) insufficient evidence error where an annotated relation can not be inferred from the corresponding document, and 3) evidence error where the sentence-level evidence of a relation is wrongly annotated. The number of errors in the three categories is 4, 44, and 90, respectively. We exhibit more details in the appendix.


%
\section{Task, Methods, and Datasets}
\subsection{Task Description}
Given a document $d$ and an entity set $\mathcal{E} = \left\{e_{i}\right\}_{i=1}^{n}$ in $d$, the target of document-level relation extraction is to predict all of the relations between entity pair $\left(e_{i}, e_{j}\right)_{i, j=1 \ldots n; i \neq j}$ among $\mathcal{R} \cup\{\mathrm{NA}\}$. $\mathcal{R}$ is the predefined relations set . $\mathrm{NA}$ indicates that there is no relation between an entity pair. $e_i$ and $e_j$ denote subject and object entities. An entity may appear many times in a document, we use set $\left\{m_{j}^{i}\right\}_{j=1}^{N_{i}}$ to distinguish the mentions of each entity. We finally build the extracted relation triples into the form of $\left\{\left(e_{i}, r_{i j}, e_{j}\right) \mid e_{i}, e_{j} \in \mathcal{E}, r_{i j} \in \mathcal{R}\right\}$.

\subsection{Methods}
We choose one of the most representative models from each category of document-level RE models (DocuNet from graph-based methods and ATLOP from transformer-based methods) to produce attributions by feature attribution (FA) methods. We choose Integrated Gradient (IG) as our attribution method due to its verified simplicity and faithfulness~\cite{sundararajan2017axiomatic}, which renders IG applicable in other text-related tasks~\cite{mudrakarta2018dida,liu2019incorporating,bastings2020elephant,hao2021self,liu2022saliency}.  
\paragraph{Integrated Gradient}
Integrated Gradient is a reference-based method that calculates both the model output on the input and that on a reference point. The difference between the outputs is distributed as an importance score for each token. Specifically, given an input $x$ and reference point $x^{\prime}$, IG computes the linear integral of the gradients $g_i$ along the $i^{th}$ dimension from $x^{\prime}$ to $x$ by,
$$
g_{i}=\left(x_{i}-x_{i}^{\prime}\right) \times \int_{\alpha=0}^{1} \frac{\partial F\left(x^{\prime}+\alpha \times\left(x-x^{\prime}\right)\right)}{\partial x_{i}} d \alpha,
$$
\noindent where $\frac{\partial F(x)}{\partial x_{i}}$ indicates the gradient of an output $F(x)$ to $x$. As set in other text-related tasks~\cite{wallace2019allennlp}, we set $x^{\prime}$ as a sequence of embedding vectors with all zero values. 

\subsection{Datasets}
\label{datasets}

\paragraph{DocRED and $\text{DocRED}_\text{Scratch}$.}
DocRED contains 56,354 human-annotated relational facts, which can be categorized into 96 relation types. Most of the relational facts ($61.1\%$) can only be identified by reasoning~\cite{yao2019docred}. Recently, \citet{huang2022does} argue that the recommend-revise scheme adopted by DocRED in annotation leads to an obvious bias toward popular entities and relations. They rectify the bias by re-annotating 96 randomly selected documents (from the validation set of DocRED) from scratch and propose $\text{DocRED}_\text{Scratch}$. The distribution of $\text{DocRED}_\text{Scratch}$ shifts largely from the training set of DocRED, which renders it applicable for testing the generalization ability of models trained on DocRED. 

\paragraph{$\text{DocRED}_\text{HWE}$}
We propose $\text{DocRED}_\text{HWE}$ with the following features: 1) $\text{DocRED}_\text{HWE}$ contains 699 documents with 27,732 evidence words (10,780 evidence phrases) annotated by humans for 7,342 relational facts among 13,716 entities. 2) We annotate 1,521 pronouns referring to different entities, which are necessary to predict corresponding relations between entity pairs and neglected in DocRED. 3) At least 3,308 out of 7,342 ($45.1\%$) relational facts require reading multiple sentences for extraction.


\section{Experiment and Analysis}
\label{experiments}

\subsection{Analyzing Decision Rules of Models}
\label{sec1}
We employ IG as our attribution technique to characterize the decision rules of models, which help us observe some potential risks in the SOTA models.

\begin{figure}
  \centering
  \includegraphics[width=1\columnwidth]{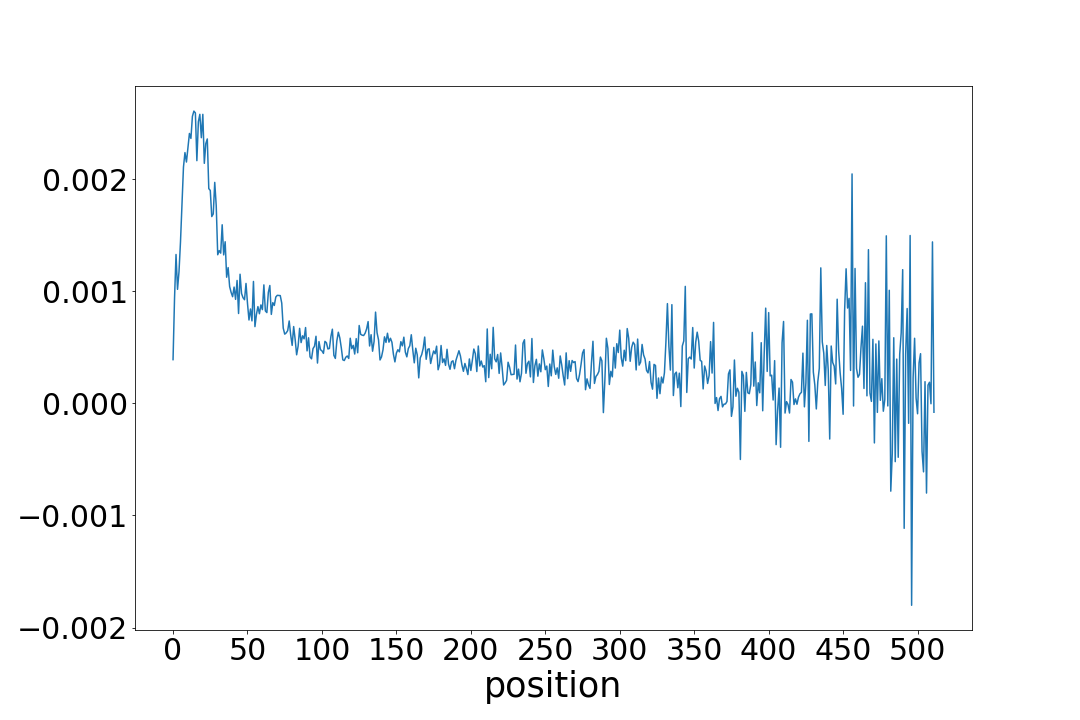}
  \caption{Mean attribution value distribution of $\text{ATLOP}_\text{RoBERTa}$ on different positions of documents in the validation set of DocRED. A similar shape of the curve emerges when attributing DocuNet. We only exhibit one of the curves due to the limited space.}
  \label{fig:sharp_ig_dis_mean_fig}
\end{figure}

\paragraph{Position Discrimination.}
After being encoded by models, each token possesses its semantic meaning (word embedding) and position information (position embedding). Before analyzing the semantic meaning, we first visualize the contribution of position information to the predictions according to the attribution values. As shown in Figure~\ref{fig:sharp_ig_dis_mean_fig}, tokens in certain positions will affect final predictions more significantly than the words in other positions. In other words, models will discriminate words according to their positions in a document, even though the annotated rationales are almost uniformly distributed across the documents. 
We posit two reasons: (1) models distort the features from positions in the process of learning and spuriously correlate certain positions with the final predictions; (2) the position embeddings are wrongly trained (unsupervised), deviating from their original function of representing the position information. Furthermore, we observe more significant variances in those positions, roughly from 450 to 500, because the number of documents that are longer than 450 is small.

Note that the learned position discrimination may happen to apply to the test set of DocRED. However, the distributional shifts in real-world applications can render the spurious pattern no longer predictive. The generalization ability of models will be severely destroyed. 

\begin{figure}
  \centering
  \includegraphics[width=1\columnwidth]{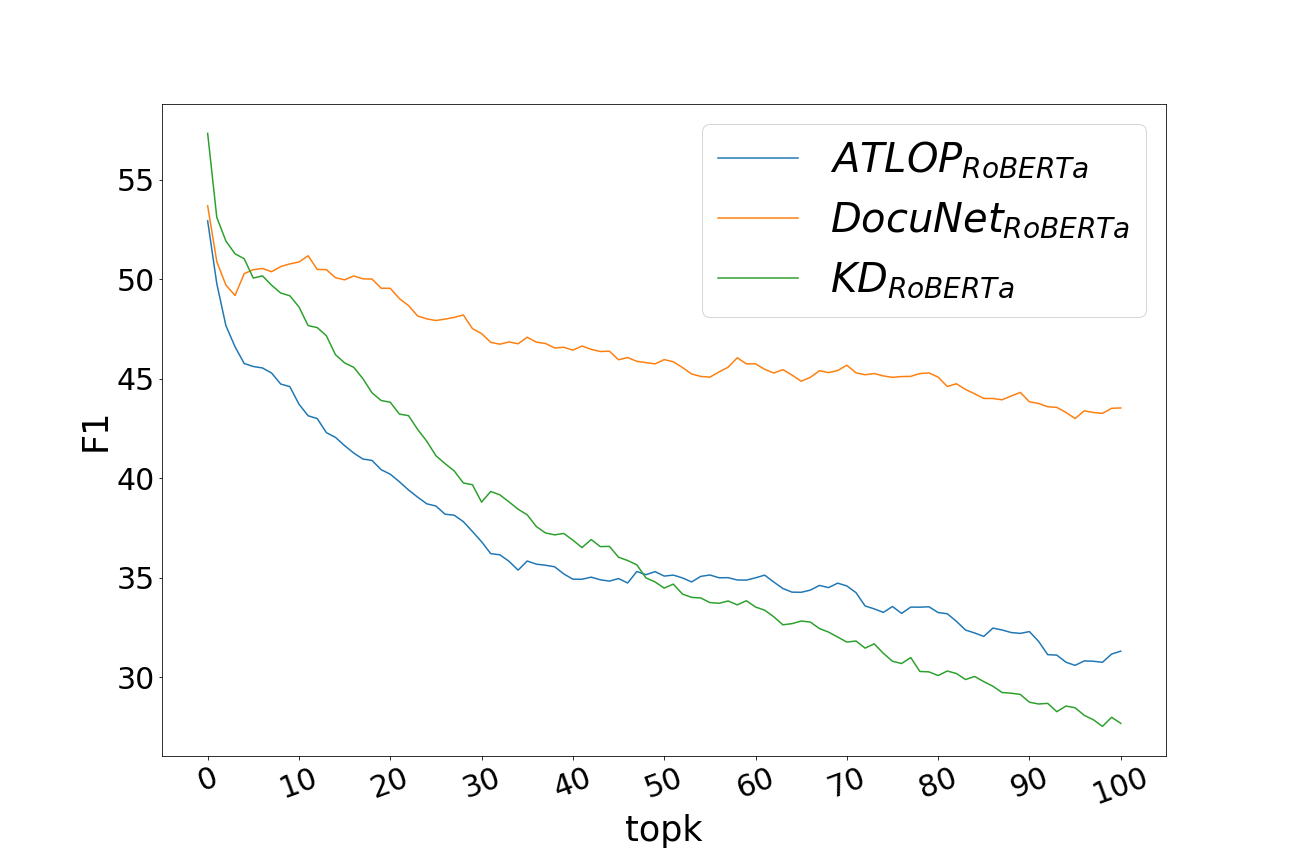}
  \caption{Performance based on top $K$ attributed words}
  \label{fig:enp_topk_fig}
\end{figure}

\paragraph{Narrow Scope of Reasoning.}


To observe the words that are necessary for a model to infer the right relations, we first investigate their number, representing the reasoning scope of models. Specifically, we design a template in the form of ``A X B'', where A and B denote the given entity pair and X can be either a word sequence or a single word. We regard X as necessary when models accurately predict the relation $r_{AB}$ between A and B according to the template. We set X to the top $K$ attributed tokens of $r_{AB}$ and the position order of the tokens is the same as that in the original document. The performance of models on the validation set of DocRED is shown in Figure~\ref{fig:enp_topk_fig}. Adding the highest attributed words surprisingly results in a performance decline. The contribution of position is significant, which is consistent with the results shown in Figure~\ref{fig:sharp_ig_dis_mean_fig}. Most importantly, we observe that models can achieve $53\%$ F1-score when only given names of entity pairs without access to the context, which remains at about $85\%$ of their original performance. Models perform reasoning in a strikingly narrow scope. If the phenomenon is reasonable, it indicates that such a few words are enough to explain rationales for the right predictions. To verify the assumption, We visualize these words in the next paragraph.


\begin{figure}
  \centering
  \includegraphics[width=1\columnwidth]{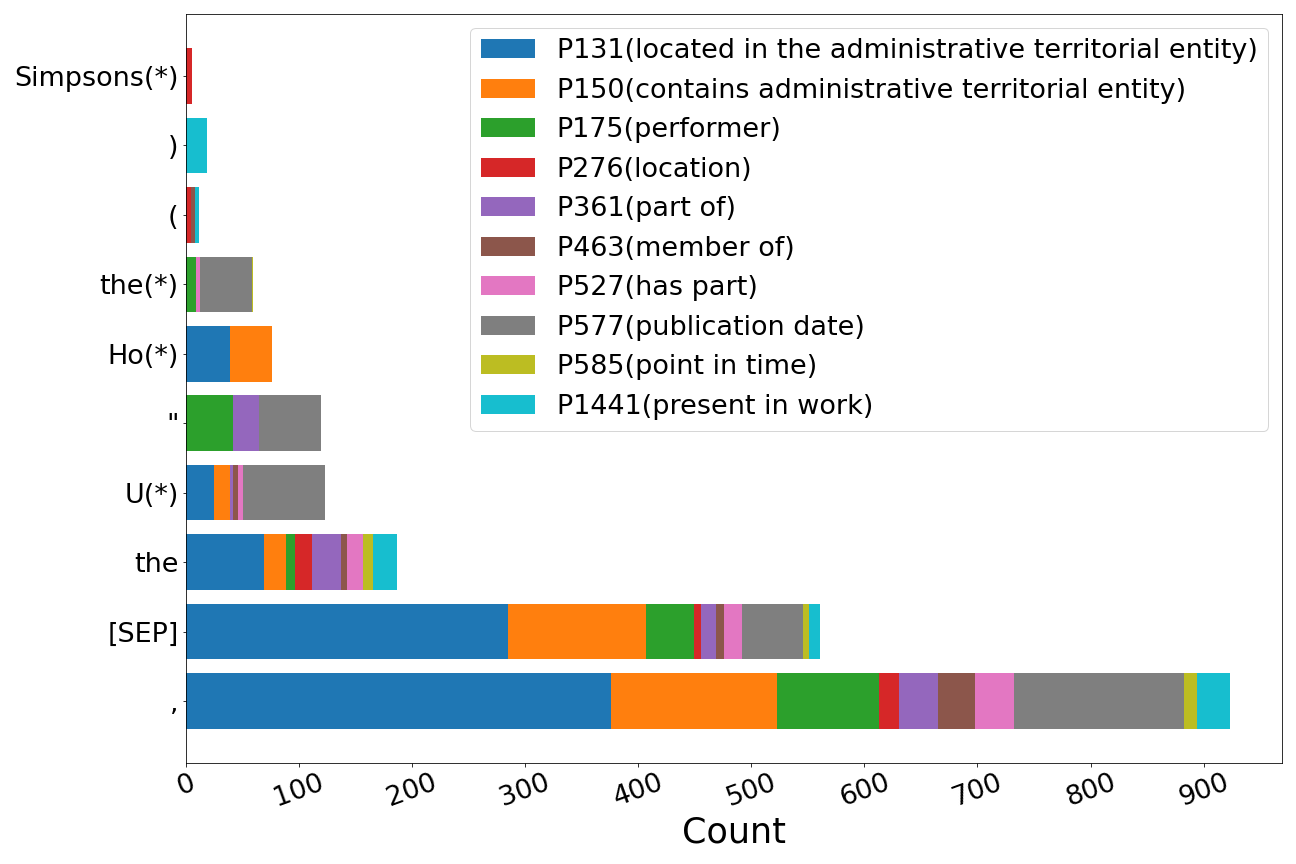}
  \caption{Statistics of word-level evidence of $\text{ATLOP}_\text{RoBERTa}$. The signal `*' denotes that the corresponding token belongs to entity names. We observe a similar phenomenon when counting the evidence words of DocuNet. We only exhibit one result due to the limited space.}
  \label{fig:ori_stat_cor_fig}
\end{figure}

\paragraph{Spurious Correlations.} 
We select the top five attributed words to visualize the evidence words of models shown in Figure~\ref{fig:ori_stat_cor_fig}. The attributions reveal that the SOTA models on DocRED largely rely on some non-causal tokens (e.g., entity name and some punctuations) to make the right predictions, which exerts a negative effect on learning the rationales. We can observe that the full-stop token, for example, plays a pivotal role in the predictions. Note that some special tokens (`[SEP]' and `[CLS]') are demonstrated to serve as ``no-op'' operators~\cite{clark2019what}. The reliance on these special tokens may not be a problem because the two tokens are guaranteed to be present and are never attacked. However, the reliance on non-causal tokens renders a model fragile, untrustworthy, and far from being deployed into real-world scenarios as non-causal tokens can easily be attacked through substitutions, paraphrasing, changes in writing style, and so on. As shown in Table~\ref{attack-table}, if models learn to predict according to non-causal tokens, then each attack in these tokens will easily be successful. This severely destroys the robustness of models. The visualization indicates that models learn abundant spurious correlations (e.g., entity names and irrelevant words) to minimize the training error.
We further prove that the spurious correlations are caused by selection bias in both pre-training and finetuning procedures. The details of the proof are given as follows.

\begin{wrapfigure}{r}{0pt}
    \centering
    \includegraphics[width=0.5\columnwidth]{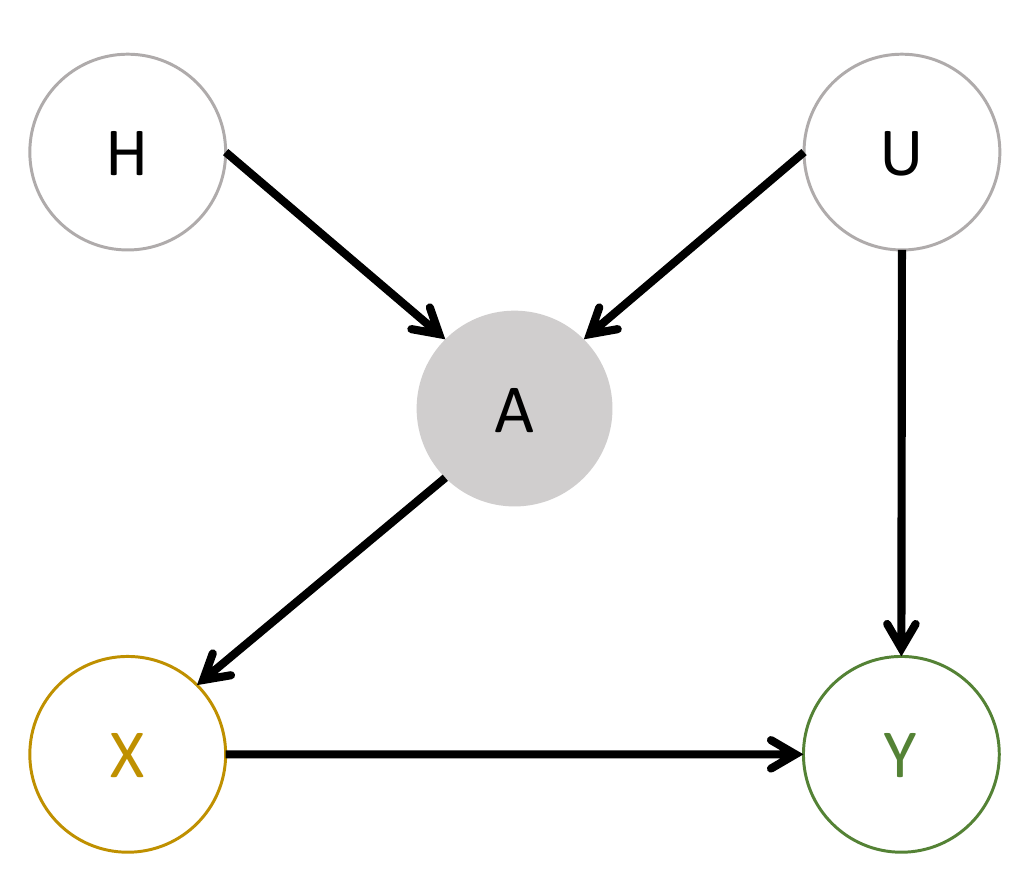}
    \caption{Causal graph of the sampling process.}
    \label{fig:dag-fig}
\end{wrapfigure}

\paragraph{Analysis of Underlying Causes.}
We shed some light on the underlying causes of learning spurious correlations. We argue that the common ground of the highly attributed non-causal tokens is that they are either high-frequency function tokens or tokens that frequently co-occur with the corresponding relations. 
Although most transformer-based pre-trained language models (PLMs) are expected to maximize the probability of current word $Y$ given its context $X$, which is represented by conditional distribution $P(Y|X)$, they have instead learned $P(Y|X, A)$, where $A$ denotes the access to the sampling process. 
The selection bias results in spurious correlations between high-frequency function tokens and current tokens. Specifically, we explain the causal relationships between variables during pre-training PLMs and represent it in a causal directed acyclic graph (DAG) as shown in Figure~\ref{fig:dag-fig}. As the high-frequency function words $H$ possess grammatical meaning (e.g.,`.' and `the'), they are more possible to be sampled either in training corpus or context, while other words $U$ are relatively less likely to access the sampling process or context. 
The phenomenon is represented by $H \rightarrow A$ and $U \rightarrow A$, where directed edges denote causal relationships between variables. However, the semantic meaning (word embedding) of the current word $Y$ largely depends on the words carrying an explicit semantic meaning, which is depicted by $U \rightarrow Y$. In linguistics, content words contribute to the meaning of sentences where they occur, and function words express grammatical relationships among other words. Their combinations, which are implicit and hard to be intervened, form natural language expressions. The process can be described by $A \rightarrow X$, where $A$ determines the word distribution over contexts.


Existing PLMs are pre-trained on a given corpus, indicating that $A$ is given. Conditioning on $A$, the unconditionally independent variables $H$ and $U$ become dependent, which is described as collider bias~\cite{pearl2009causality}. Due to the causal relationship between $U$ and $Y$, $H$ and $Y$ are spuriously correlated. That is to say, models tend to spuriously correlate high-frequency function words with any word, including word-level evidence which causes relations. Therefore, spurious correlations between high-frequency function words and relations are learned by models and represented in Figure~\ref{fig:ori_stat_cor_fig}. 

Meanwhile, we can also observe spurious correlations between entity names and relations. Our analysis of the underlying causes is roughly the same as we mentioned before. We regard $H$ as high-frequency entities from the relation-specific documents in Wikipedia, $U$ as evidence words that causally determine the relation, $Y$ as predictions, and $X$ as documents. Given $X$ and $A$, models learn spurious correlations between $H$ and $Y$.




\begin{table*}
    \centering
    \scalebox{0.7}{
    \begin{tabular}{l|ccccc|cccccc}
    \toprule
        \multirow{2}{*}{\textbf{Model}} & \multicolumn{2}{c}{\textbf{Mask.}} & \textbf{ASA} & \multicolumn{2}{c}{\textbf{SSA}} & {\textbf{EM}} & {\textbf{ES}} &  {\textbf{ER}} & {\textbf{Val}} & {\textbf{HWE}} & {\textbf{Scratch}} \\ \cline{2-12} 
        
        & P2N & UP  & UP &  P2N & UP & F1  & F1 & F1 &  F1 & F1 & F1 \\\hline
        
         $\text{ATLOP}_\text{BERT}$~\cite{zhou2020documentlevel}
         & \bm{$20.21$} & \bm{$79.43$} & \bm{$90.38$} & $\ \ 6.47$ & $93.46$ & $\ \ 6.39$ & $\ \ 6.08$ & $14.16$ & $61.09$ & $57.69$ & $40.56$\\ 
        $\text{ATLOP}_\text{RoBERTa}$~\cite{zhou2020documentlevel}
         & $16.51$ & $82.98$  & $90.42$ & $\ \ 3.85$ & $96.02$ & $27.29$ & $\ \ 7.35$ & $17.50$ & $63.18$ & $58.43$ & $42.12$ \\
        $\text{DocuNet}_\text{RoBERTa}$~\cite{zhang2021document}
         & $16.49$ & $83.19$  & $91.48$ & $\ \ 2.82$ & $97.17$ & $\ \ 8.62$ & $\ \ 8.08$ & $18.55$ & $63.91$ & $59.58$ & $42.78$ \\
         $\text{SSAN}_\text{RoBERTa}$~\cite{xu2021entity}
         & $13.68$ & $85.48$ & $91.23$ & $\ \ 1.73$ & $98.26$ & $35.41$ & $\ \ 6.09$ & $22.72$ & $62.08$ & $58.37$ & $48.74$ \\ 
        $\text{EIDER}_\text{RoBERTa}$~\cite{xie2022eider}
        & $14.24$ & $85.36$ &  $92.78$ & $\ \ 2.12$ & $97.88$ & \bm{$35.45$} & \bm{$\ \ 8.46$} & \bm{$23.00$} & $64.28$ & {$60.62$} & \bm{$49.95$} \\
        $\text{KD}_\text{RoBERTa}^\dagger$~\cite{tan2022documentlevela}
         & $10.77$ & $88.69$ & $95.46$ & \bm{$\ \ 1.28$} & \bm{$98.72$} & $29.74$ & $\ \ 7.57$ & $20.35$ & \bm{$67.12$} & \bm{$62.87$} & $45.82$ \\
        
    \bottomrule
    \end{tabular}}
    \caption{\label{attack-table} Results of different attacks. Model denoted by $\dagger$ is trained by extensive distantly supervised data. Val, HWE, and Scratch denote the validation set of DocRED, $\text{DocRED}_\text{HWE}$, and $\text{DocRED}_\text{Scratch}$. To observe the ratios of changed predictions in various attacks based on human-annotated word-level evidence, we propose P2N and UP. They denote the ratio of ``negative predictions changed from positive predictions'' to ``original positive predictions'', and ``unchanged positive predictions'' to ``original positive predictions'', respectively. }
\end{table*}

\subsection{Attacks on the SOTA DocRE Models} 
\label{sec2}
In this section, we propose several RE-specific attacks to reveal the following facts: (1) The decision rules of models are largely different from that of humans. (2) Such a difference will severely damage the robustness and generalization ability of models: If a certain model always neglects the rationales in DocRE, it can hardly be aware of the tiny but crucial modifications on rationales. We introduce more details of our proposed attacks as follows.
\paragraph{Word-level Evidence Attacks.}
We present three kinds of attacks according to our proposed word-level evidence annotation: (1) \textit{Masked word-level evidence attack} where all of the human-annotated word-level evidence (HWE) is directly masked; (2) \textit{Antonym substitution attack (ASA)} where a word in HWE is replaced by its antonyms; (3) \textit{Synonym substitution attack (SSA)} where a word in HWE is replaced by its synonyms. Since some evidence words do not have antonyms or synonyms in WordNet~\cite{miller1995wordnet}, we attack the rest of the words in HWE. Note that we only attack the HWE of those relation facts that have a single reasoning path to make sure our antonym/synonym substitution will definitely change/keep the original label. Specifically, in ASA, we first select the first suitable word in HWE that either possesses its antonym in WordNet or belongs to different forms of the verb ``be''. We generate the opposite meaning either by adding ``not'' after the ``be'' verbs or substituting the word with its antonym. In SSA, the first suitable word in HWE will be replaced by its synonyms. We conduct ASA and SSA on 2,002 and 5,321 relational facts, respectively.

The results of the three kinds of attacks are shown in Table~\ref{attack-table}. Under the masked word-level evidence attacks, the evidence supporting the relational facts is removed. The relations between entity pairs are supposed not to exist. However, we can observe that, as to the best performance, no more than $21\%$ of predictions is even changed. Models still predict the same relations even if they are erased, which leads to at least a $79\%$ decline in the performance of models. As to ASA, the semantic meanings of evidence are changed to the opposite. Models are expected to alter their predictions. However, the SOTA models alter no more than $10\%$ predictions after the attack, which indicates that the performance of models will sharply drop by at least $90\%$ under ASA. The results of SSA are roughly the same as ASA. According to the experimental results of previous attacks, we can attribute the good performance of models under SSA to the fact that models are hardly aware of rationales. All three kinds of attacks confirm the conclusion that the decision rules of models are largely different from that of humans. The difference severely damages the robustness of models. 

\paragraph{Entity Name Attacks} As shown in Section~\ref{sec1}, we observe that models rely largely on tokens in entities. To further investigate the extent to which models depend on entity names to improve their performance, we design a few attacks to exhibit their bottleneck. We propose (1) mask entity attack (EM) where we directly mask all entity names, (2) randomly shuffled entity attack (ER) where we randomly permute the names of entities in each document, and (3) out-of-distribution (OOD) entity substitution attack (ES) where we use entity names that have never occurred in training data to substitute the entity names in an input document. As shown in Table~\ref{attack-table}, we observe significant declines in the F1 scores from all models. The experimental results are shown in Table~\ref{attack-table}. The most significant performance decline occurs when attacking $\text{KD}_\text{RoBERTa}$ by ES, where the F1-score drops from $67.12\%$ to $7.57\%$.

The results of entity name attacks show that models spuriously correlate entity names with the final predictions. In other words, they make predictions according to entity names. The poorer the performance, the more spurious correlations are learned. The differences are: (1) EM removes original entity name information to detect spurious correlations;
(2) ER modifies original entity name information to attack the learned spurious correlations, making them misleading to further test the robustness of models;
(3) OOD-ES removes original entity name information and introduces new OOD entity name information, evaluating the generalization ability of models on tackling the unseen entity name information without the help of spurious correlations.


\subsection{Evaluation Metric}
\label{sec3}
In Section~\ref{sec2}, we demonstrate that the decision rules of models should approach that of humans to improve the understanding and reasoning capabilities of models. The desiderata of the capabilities and the similar conclusions are also presented in other NLP tasks~\cite{jia2017adversariala,wang2022identifying}. However, how do we measure the extent to which models possess these capabilities? In other words, how to measure the distance between the decision rules of models and that of humans? In previous work, they calculate F1-score over the evidence sentences. Models are trained to recognize the corresponding right evidence sentences when they extract a relational fact. Despite the plausible process, the recognized holistic evidence sentences fail to provide fine-grained word-level evidence, resulting in unfaithful observations discussed in Section~\ref{motivations}.
Furthermore, models' performance of predicting evidence sentences can not represent their understanding and reasoning capabilities: the blackbox process of learning how to predict evidence may introduce other new problems in the newly learned decision rules.


\begin{figure}
  \centering
  \includegraphics[width=1\columnwidth]{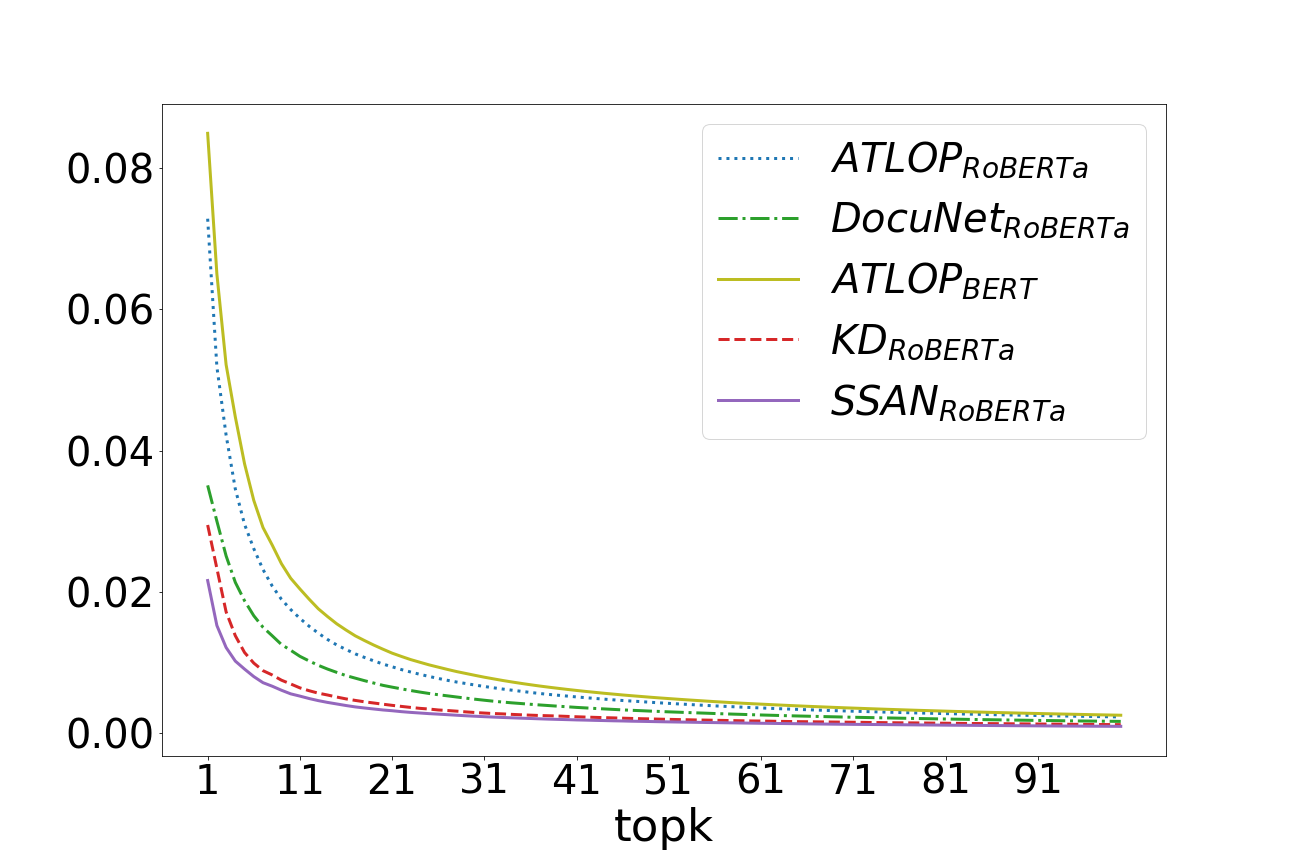}
  \caption{MAP curves of different models.}
  \label{fig:nmap_fig}
\end{figure}




To solve the issue, we introduce mean average precision (MAP)~\cite{zhu2004recall} to evaluate the performance of models and explain their reliability. We also visualize the MAP values of the models.

MAP is a widely adopted metric to evaluate the performance of models, including Faster R-CNN~\cite{ren2015fasterrcnn}, YOLO~\cite{redmon2016you}, and recommender systems~\cite{ma2016user}. We note that evaluating recommender systems and measuring the capabilities of models share a common background. Intuitively, we can consider ``the human-annotated evidence words'' as ``the relevant items for a user'', and ``the most crucial words considered by a certain model'' as ``the recommended items of a recommender system''. Consequently, given top $K$ words with the highest attribution values, the formula of MAP over $T$ relational facts can be written by, 
\begin{equation}
\small
\mathrm{MAP}(K)=\frac{1}{T} \sum_{t=1}^{T} \mathrm{AP}_{t}(K)=\frac{1}{T} \sum_{t=1}^{T} \frac{1}{K} \sum_{i=1}^{K} P_{t}(i) \cdot \mathbf{1}_{t}(i),
\end{equation}
where $\mathbf{1}_t(i)$ denotes the indicator function of the $i$-th important word for predicting the $t$-th relational fact. The output value of $\mathbf{1}_t(i)$ equals $1$ if the word is in the human-annotated word-level evidence. Else it equals $0$. The selection of K, similar to the evaluation metrics in recommender systems, depends on the demand of RE practitioners and is often set to 1, 10, 50, and 100. Also, we can select all the possible values of K to form a MAP curve and measure the AUC to holistically evaluate the understanding ability of models. For each relational fact, words ``recommended'' by models will be evaluated according to 1) how precise they perform the human-annotated word-level evidence, and 2) the ``recommending'' order of these important words determined by their attribution values. Based on MAP, we measure the extent to which the decision rules of models differ from that of humans. Due to the mechanism of EIDER where documents and the predicted sentences from documents are combined together to predict by truncation, it is impractical to attribute EIDER by gradient-based methods. We compute MAP for other SOTA models. The results are shown in Figure~\ref{fig:nmap_fig}. We can observe that the MAP values of SOTA models are all below $10\%$, which is far less than the average level of normal recommender systems. Obviously, existing models fail to understand the documents as humans do, which explains the reason why they are vulnerable to our proposed attacks. 

In this section, we use MAP to evaluate to which extent a model makes decisions like a human, which indicates the brittleness and robustness of a model. Models can explore many ways to achieve a good performance on the test set (represented by F1 score), including greedily absorbing all correlations found in data, recognizing some spurious patterns, etc., but MAP will tell us which model is trustworthy or robust and can be deployed in real-world applications.

\subsection{Discussion}
\label{sec4}
In this section, we discuss the connections between some experimental results to give some instructive advice. 
First, we can observe that for the models whose MAP value is larger, their performance under word-level evidence-based attacks will be better. MAP curve reflects the extent to which models possess understanding and reasoning abilities. As shown in Figure~\ref{fig:nmap_fig}, the various extents can be described from high to low by $\text{ATLOP}_\text{BERT}>\text{ATLOP}_\text{RoBERTa} >\text{DocuNet}_\text{RoBERTa}>\text{KD}_\text{RoBERTa} \approx \text{SSAN}_\text{RoBERTa} $, which is consistent with the performance levels under mask word-level evidence attack and antonym substitution attack represented from high to low by $ \text{ATLOP}_\text{BERT} > \text{ATLOP}_\text{RoBERTa} > \text{DocuNet}_\text{RoBERTa} > \text{KD}_\text{RoBERTa} \approx \text{SSAN}_\text{RoBERTa}$. Furthermore, if the decision rules of models largely differ from that of humans (MAP value is small), it will be ambiguous to identify which kind of attack the models will be vulnerable to. According to the results in Table~\ref{attack-table}, the performance of models are irregular under entity name attacks. The underlying causes can be any factors that can influence the training effect on a model.

Although training on extensive distantly supervised data can lead to the performance gain on the validation set of DocRED and $\text{DocRED}_\text{HWE}$, it also renders the poor understanding and reasoning capabilities of $\text{KD}_\text{RoBERTa}$ according to Figure~\ref{fig:nmap_fig}, which makes it be the most vulnerable model under mask word-level evidence attack and antonym substitution attack. As shown in Table~\ref{attack-table}, the generalization ability of $\text{KD}_\text{RoBERTa}$ is also weakened when compared with $\text{EIDER}_\text{RoBERTa}$ on $\text{DocRED}_\text{Scratch}$, which does not use any extra training data and predicts through evidence sentences annotated by humans. $\text{EIDER}_\text{RoBERTa}$ simultaneously enhances the performance, generalization ability, and robustness of models. We can observe its stronger robustness under entity name attacks, outstanding performance on the validation set of DocRED and $\text{DocRED}_\text{HWE}$, and stronger generalization ability on $\text{DocRED}_\text{Scratch}$. The success of $\text{EIDER}_\text{RoBERTa}$ indicates that rationales considered by humans are of the essence in DocRE.

All the results indicate that guiding a model to learn to predict by the evidence of humans can be the essential way to improve the robustness of models, thus making models trustworthy in real-world applications.

\section{Limitation}
In this paper, we propose $\text{DocRED}_\text{HWE}$ and introduce a new metric to select the most robust and trustworthy model from those well-performed ones in DocRE. However, all data in DocRED are sampled from Wikipedia and Wikidata, which indicates that training and test data in DocRED can be identically and independently distributed (i.i.d. assumption). The i.i.d. assumption impedes our demonstration of the intuition: A model with a higher MAP will obtain a higher F1 score on the test set. Due to the i.i.d. assumption, models can succeed in obtaining a higher F1 score by greedily absorbing all correlations (including spurious correlations) in the training data. To strictly demonstrate the intuition, we need a test set that exhibits different and unknown testing distributions. In addition, expanding the research scope to a cleaner Re-DocRED and analyzing the role of unobservable wrong labels are also crucial and interesting ideas. We leave them as our future work. 
\section{Conclusion}
Based on our analysis of the decision rules of existing models on DocRE and our annotated word-level evidence, we expose the bottleneck of the existing models by our introduced MAP and our proposed RE-specific attacks. We also extract some instructive suggestions by exploring the connections between the experimental results.

We appeal to future research to take understanding and reasoning capabilities into consideration when evaluating a model and then guide models to learn evidence from humans. Based on proper evaluation and guidance, significant development can be brought to the document-level RE, where the performance, generalization ability, and robustness of models are more likely to be improved.

\section*{Acknowledgement}
This work is partially supported by funds from Arcplus Group PLC (Shanghai Stock Exchange: 600629).

\bibliography{custom}
\bibliographystyle{acl_natbib}

\appendix

\section{Details of Attacks}
\begin{figure}
  \centering
  \includegraphics[width=1\columnwidth]{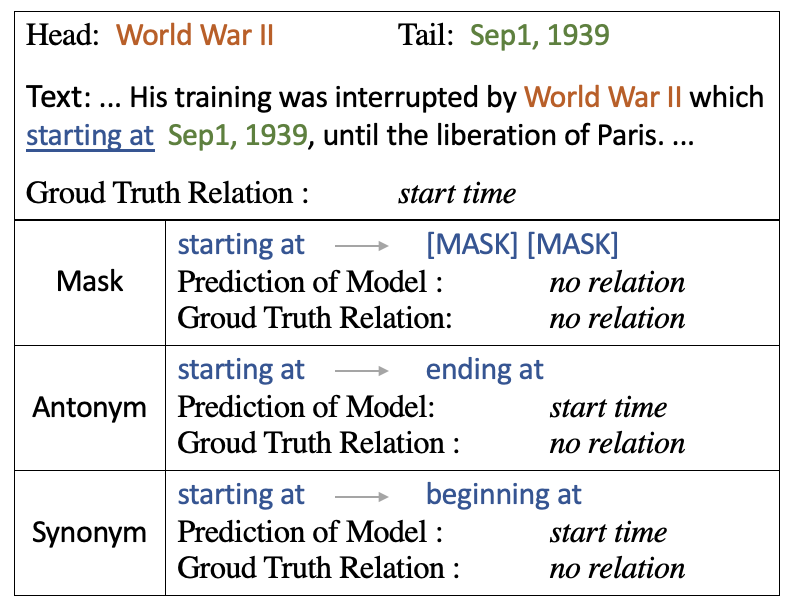}
  \caption{An example for the three kinds of attacks.}
  \label{fig:attack-example}
\end{figure}

We give an example to illustrate our proposed three kinds of word-level evidence attacks. The example is shown in Figure~\ref{fig:attack-example}

\section{Annotation Errors in DocRED}
We provide the details of all our corrected errors in our selected 699 documents from the validation set of the original DocRED. All the error descriptions are shown in Table~\ref{error1}, Table~\ref{error2}, and Table~\ref{error3}. Annotators correct three kinds of annotation errors, which are exhibited in Table~\ref{error4} and Table~\ref{error5}. ``Err.1'' denotes relation type error where annotators wrongly annotate a relation type between an entity pair. ``Err.2'' denotes insufficient evidence error where an annotated relation can not be inferred from the corresponding document. ``Err.3'' denotes evidence error where the sentence-level evidence of a relation is wrongly annotated. 

\begin{table*}
    \centering
    \scalebox{0.8}{
    \begin{tabular}{l|c|p{11cm}<{\centering}}
    \toprule
        Document Title & Rel. & Error Description \\ \hline
        The Time of the Doves & 2 & The relation can only be inferred by the first sentence. \\ 
        The Time of the Doves & 4 & The relation is not P150. \\ 
        Hélé Béji & 8 & No evidence can be found for this relation. \\  
        Hélé Béji & 1 & We can't infer relation P569 from the first evidence sentence. \\  
        Ne crois pas & 1 & The only evidence sentence of P27 is the sentence 5 instead of 0. \\  
        Ne crois pas & 9 & The only evidence sentence of P27 is the sentence 5 instead of 2. \\  
        Ne crois pas & 14 & The only evidence sentence of P1344 is the sentence 7 instead of 2. \\  
        Asian Games & 5 & No evidence can be found for this relation. \\  
        Asian Games & 7 & No evidence can be found for this relation. \\  
        The Longest Daycare & 10 & The second sentence does not clearly indicate that David is the director and only the third sentence indicates it, so the evidence is [0,3] \\  
        The Longest Daycare & 28 & the zeroth sentence can't infer that Simpsons are from the United States and Only the seventh sentence indicates it, so the evidence is [0,7] \\  
        South Gondar Zone & 1 & P150 can not be inferred, evidence is null, can't find evidence \\  
        South Gondar Zone & 4 & P17 can not be inferred according to the given document. \\  
        South Gondar Zone & 16 & P403 can not be inferred according to the given document. \\  
        South Gondar Zone & 3 & Evidence of the third relation(P27) in labels is [0,1] instead of
[0,1,2] \\  
        Milton Friedman ... & 1 & "Evidence of the first relation(P31) in labels is [0] instead of
[0,3]" \\  
        Milton Friedman ... & 8 & Evidence of the eighth relation(P108) in labels is [8] instead of [7,8] \\  
        Fedor Ozep & 2 & Evidence of the second relation(P20) in labels is [6] instead of [0,6] \\  
        TY.O & 1 & Evidence of P264 is [1] instead of [0,1] \\  
        TY.O & 3 & Evidence of P175 is [3] instead of [0, 3, 4] \\  
        TY.O & 10 & Evidence of P175 is [0] instead of [0, 4] \\  
        TY.O & 13 & Evidence of 162 is [0,3] instead of [0, 3, 4] \\  
        TY.O & 14 & Evidence of P175 is [0,3] instead of [0, 3, 4] \\  
        TY.O & 20 & Evidence of P175 is [0,3] instead of [0, 3, 4] \\  
        TY.O & 29 & Evidence of P175 is [0,3] instead of [0, 3, 4] \\  
        Front of Islamic ... & 1 & Evidence of P1412 is [0,2] instead of [0, 2, 4] \\ 
        Front of Islamic ... & 2 & Evidence of P1412 is [0,2] instead of [0, 2, 4] \\  
        Front of Islamic ... & 3 & Evidence of P37 is [0,3] instead of [0, 3, 4] \\  
        Front of Islamic ... & 4 & Evidence of P1412 is [0,2] instead of [0, 2, 4] \\  
        Front of Islamic ... & 5 & Evidence of P1412 is [0,3] instead of [0, 3, 4] \\  
        Rufus Carter & 7 & P131 represents "located in the administrative territorial entity",but it can not be inferred according to the given document. \\  
        Rufus Carter & 8 & P150 can not be inferred according to the given document. \\  
        Smoke Break & 1 & Evidence of P577 is [1] instead of [1,8] \\  
        Smoke Break & 2 & Evidence of P264 is [1] instead of [1,2] \\  
        Smoke Break & 3 & Evidence of P162 is [2] instead of [0,2] \\  
        Bambi II & 6 & P17 can not be inferred according to the given document. \\  
        Bambi II & 8 & P272 can not be inferred according to the given document. \\  
        Bambi II & 13 & P272 can not be inferred according to the given document. \\  
        Bambi II & 15 & P272 can not be inferred according to the given document. \\  
        Assassin's Creed Unity & 1 & P178 can not be inferred according to the given document. \\  
        Assassin's Creed Unity & 1 & P178 can not be inferred according to the given document. \\  
        Assassin's Creed Unity & 16 & P577 can not be inferred according to the given document. \\  
        Assassin's Creed Unity & 13 & P179 can not be inferred according to the given document. \\  
        Assassin's Creed Unity & 3 & P123 can not be inferred according to the given document. \\  
        Mehmet Çetingöz & 1 & P17 can not be inferred according to the given document. \\  
        Mehmet Çetingöz & 2 & P17 can not be inferred according to the given document. \\  
        Mehmet Çetingöz & 9 & P17 can not be inferred according to the given document. \\  
        Mehmet Çetingöz & 12 & P17 can not be inferred according to the given document. \\  
        Mehmet Çetingöz & 10 & P17 can not be inferred according to the given document. \\  
        Baltimore and ... & ~ & Evidence of P17 is [0,2,3] instead of [0,2] \\  
        Baltimore and ... & ~ & Evidence of P279 is [2] instead of [2,4] \\  
        Dante Alighieri Society & 6 & Evidence of P571 is [0,2] instead of [0,2,5] \\  
        Osaka Bay & 25 & Evidence of P17 is [0,3,11] instead of [0,11] \\  
        Osaka Bay & 36 & Evidence of P17 is [0,3,11] instead of [0,11] \\  
        Osaka Bay & 38 & Evidence of P17 is [0,3,11] instead of [0,11] \\  
        Liang Congjie & 8 & relation can not be inferred from the context \\  
        \bottomrule
    \end{tabular}}
    \caption{Wrong annotations in the original DocRED.}
    \label{error1}
\end{table*}

\begin{table*}
    \centering
    \scalebox{0.8}{
    \begin{tabular}{p{4cm}|c|p{11cm}<{\centering}}
    \toprule
        Document Title & Rel. & Error Description \\ \hline
        University (album) & 18 & Evidence of P264 should be [3,4] \\  
        University (album) & 19 & Evidence of P175 should be [3] \\  
        University (album) & 24 & Evidence of P527 should be [3] \\  
        University (album) & 25 & Evidence of P475 should be [0] \\  
        Lappeenranta & 2 & Evidence of P131 is [1,2,3] instead of [1,3] \\  
        Lappeenranta & 12 & Evidence of P17 is [0,2,4,5,7,9] instead of [0,2,4,7,9] \\  
        Lappeenranta & 13 & Evidence of P131 is [1] instead of [0,1] \\  
        Lappeenranta & 18 & Evidence of P131 is [1,3] instead of [1,2,3] \\  
        Ali Abdullah Ahmed & 4 & Evidence of P3373 is [6] instead of [3,6] \\  
        Ali Abdullah Ahmed & 8 & Evidence of P3373 is [6] instead of [3,6] \\  
        Ali Abdullah Ahmed & 9 & Evidence of P570 is [7] instead of [6,7] \\  
        Joseph R. Anderson & 9 & P571 can not be inferred according to the given document. \\  
        Ramblin' on My Mind & 1 & Evidence of P175 is [5] instead of [0,2] \\  
        Ramblin' on My Mind & 2 & P86 can not be inferred according to the given document. \\  
        Christopher Franke & 3 & Evidence of P463 is [1,3,4,5] instead of [0,1,3,5] \\  
        Christopher Franke & 4 & P159 can not be inferred according to the given document. \\  
        Christopher Franke & 5 & P577 can not be inferred according to the given document. \\  
        Statue of Jan Smuts & 3 & Evidence of P27 is [5] instead of [4,5] \\  
        Statue of Jan Smuts & 4 & Evidence of P27 is [5] instead of [4,5] \\  
        Robert Taylor & 1 & Evidence of P108 is [1] instead of [0,1] \\  
        Robert Taylor & 2 & Evidence of P27 is [2] instead of [4,5] \\  
        Robert Taylor & 3 & Evidence of P27 is [3] instead of [4,5] \\  
        Robert Taylor & 4 & Evidence of P27 is [4] instead of [4,5] \\  
        Sycamore Canyon & 1 & P17 can not be inferred according to the given document. \\  
        Amos Hochstein & 9 & P194 can not be inferred according to the given document. \\  
        Paul Pfeifer & 3 & P69 can not be inferred according to the given document. \\  
        Mega Man Zero & 8 & P155 can not be inferred, Virtual Console is Wii U \\  
        Soldier (song) & 1 & Evidence of P577 is [1] instead of [0,1] \\  
        Soldier (song) & 3 & Evidence of P495 is [2] instead of [0,2] \\  
        Gloria Estefan Albums Discography & 4 & P156 can not be inferred. Let It Loose and Cuts Both Ways are two albums published one after another instead of two songs from an album. They are independent of each other. There is no evidence in the context. \\  
        Anthony G. Brown & 3 & Evidence of P27 is [0,4] instead of [0,3]. \\  
        Harbour Esplanade & 3 & P17 can not be inferred according to the given document. \\  
        Harbour Esplanade & 5 & P17 can not be inferred according to the given document. \\  
        Harbour Esplanade & 6 & P17 can not be inferred according to the given document. \\  
        Henri de Buade & 3 & The relation between France and New France is colony instead of P495. \\  
        The Reverent Wooing of Archibald & 5 & P577 should be P580. \\  
        This Little Girl of Mine & 6 & The third  sentence should be removed from the evidence of P136. \\  
        This Little Girl of Mine & 9 & The zeroth sentence should be removed from the evidence of P175. \\  
        This Little Girl of Mine & 13 & The zeroth sentence should be removed from the evidence of P175. \\  
        This Little Girl of Mine & 15 & The zeroth sentence should be removed from the evidence of P264,
it only refers to the name of the head entity. \\  
        This Little Girl of Mine & 19 & The zeroth sentence should be removed from the evidence of P175
it only refers to the name of the tail entity. \\  
        This Little Girl of Mine & 20 & "The zeroth sentence should be removed from the evidence of P175,
it only refers to the name of the performer and can't infer the
relation between two sides." \\  
        Ali Akbar Moradi & 1 & Evidence of P569 should be [0]. \\  
        Ali Akbar Moradi & 2 & The zeroth sentence should be removed from the evidence of P19,
it only refers to the name of the head entity. \\  
        Ali Akbar Moradi & 3 & The zeroth sentence should be removed from the evidence of P27,
it only refers to the name of the head entity. \\  
        I Knew You Were Trouble & 2 & The zeroth sentence should be removed from the evidence of P264,
because no words related to two entities can be found in it. \\  
        I Knew You Were Trouble & 4 & The zeroth sentence should be removed from the evidence of P175
it only refers to the name of the head entity. \\  
        I Knew You Were Trouble & 5 & The zeroth sentence should be removed from the evidence of P577
it only refers to the name of the head entity. \\   
        \bottomrule
    \end{tabular}}
    \caption{Wrong annotations in the original DocRED.}
    \label{error2}
\end{table*}

\begin{table*}
    \centering
    \scalebox{0.8}{
    \begin{tabular}{p{4cm}|c|p{11cm}<{\centering}}
    \toprule
        Document Title & Rel. & Error Description \\ \hline
        I Knew You Were Trouble & 6 & The zeroth sentence should be removed from the evidence of P495
it only refers to the name of the head entity. \\  
        I Knew You Were Trouble & 7 & The zeroth sentence should be removed from the evidence of P264
it only refers to the name of the head entity. \\  
        I Knew You Were Trouble & 8 & The zeroth sentence should be removed from the evidence of P162
it only refers to the name of the head entity. \\  
        I Knew You Were Trouble & 9 & The zeroth sentence should be removed from the evidence of P361
it only refers to the name of the head entity. \\  
        Mohammed Abdel Wahab & 6 & P86 can not be inferred according to the given document. \\  
        Mohammed Abdel Wahab & 8 & P86 can not be inferred according to the given document. \\  
        Mohammed Abdel Wahab & 10 & P86 can not be inferred according to the given document. \\  
        Elbląg County & 5 & Evidence of P150 is [0,2] instead of [0,2,3]. \\  
        The Crazy World of Arthur Brown (album) & 1 & P264 represents ``brand and trademark associated with the marketing
of subject music recordings and music videos'', but here the head entity
is the same name as music, instead of a music album. \\  
        The Crazy World of Arthur Brown (album) & 6 & P264 represents ``brand and trademark associated with the marketing
of subject music recordings and music videos'', but here the head entity
is the same name as music, instead of a music album." \\  
        The Crazy World of Arthur Brown (album) & 7 & P264 represents ``brand and trademark associated with the marketing
of subject music recordings and music videos'', but here the head entity
is the same name as music, instead of a music album. \\  
        The Crazy World of Arthur Brown (album) & 8 & P264 represents ``brand and trademark associated with the marketing
of subject music recordings and music videos'', but here the head entity
is the same name as music, instead of a music album. \\  
        The Crazy World of Arthur Brown (album) & 9 & P264 represents ``brand and trademark associated with the marketing
of subject music recordings and music videos'', but here the head entity
is the same name as music, instead of a music album. \\  
        Flag of Prussia & 1 & Evidence of P155 is [0] instead of [2,4]. \\  
        Flag of Prussia & 3 & P155 should be P6. \\  
        Flag of Prussia & 7 & Evidence of P156 is [0] instead of [2,4]. \\  
        Flag of Prussia & 11 & P156 represents ``immediately following item in a series of which the subject is a part'', but here both entities are the same. \\  
        John Alexander Boyd & 11 & Evidence of P17 is [0] instead of [0,5]. \\  
        John Alexander Boyd & 12 & Evidence of P17 is [0] instead of [0,5,6]. \\  
        Municipal elections in Canada & 5 & Evidence of P17 is [8] instead of [8,11]. \\  
        Municipal elections in Canada & 7 & Evidence of P131 is [11] instead of [0,8,11]. \\  
        House of Angels & 7 & Evidence of P495 is [0,8] instead of [0,6,8]. \\  
        William James Wallace & 7 & Evidence of P17 is [0,1] instead of [1,3]. \\  
        William James Wallace & 8 & P17 can not be inferred according to the given document. \\  
        William James Wallace & 10 & P27 can not be inferred according to the given document. \\  
        William James Wallace & 11 & P17 can not be inferred according to the given document. \\  
        Black Mirror (song) & 7 & Evidence of P264 is [2] instead of [0,2]. \\  
        Michael Claassens & 5 & Evidence of P264 is [4] instead of [0,4]. \\  
        Michael Claassens & 12 & Evidence of P264 is [6] instead of [0,6]. \\  
        Lark Force & 13 & the zeroth sentence can't infer that HMAT Zealandia is from Australia. \\  
        Washington Place (West Virginia) & 9 & the zeroth sentence can't infer that Annie Washington is from the United States. \\  
        Battle of Chiari & 2 & Evidence of P276 is [0,2] instead of [0,3]. \\  
        Battle of Chiari & 6 & Evidence of P607 is [1,2] instead of [1]. \\  
        Woodlawn, Baltimore County, Maryland & 18 & Evidence of P131 is [0,5] instead of [0,4,5]. \\  
        Wagner–Rogers Bill & 1 & Evidence of P27 is [0,1] instead of [0]. \\  
        \bottomrule
    \end{tabular}}
    \caption{Wrong annotations in the original DocRED.}
    \label{error3}
\end{table*}

\begin{table}
    \centering
    \scalebox{0.65}{
    \begin{tabular}{l|c|c|c|c}
    \toprule
        Document Title & Rel. & Err. 1 & Err. 2 & Err. 3 \\ \hline
        The Time of the Doves & 2 & ~ & ~ & $\checkmark$ \\ 
        The Time of the Doves & 4 & $\checkmark$ & ~ & ~ \\ 
        Hélé Béji & 8 & ~ & $\checkmark$ & ~ \\  
        Hélé Béji & 1 & ~ & ~ & $\checkmark$ \\  
        Ne crois pas & 1 & ~ & ~ & $\checkmark$ \\  
        Ne crois pas & 9 & ~ & ~ & $\checkmark$ \\  
        Ne crois pas & 14 & ~ & ~ & $\checkmark$ \\  
        Asian Games & 5 & ~ & ~ & $\checkmark$ \\  
        Asian Games & 7 & ~ & ~ & $\checkmark$ \\  
        The Longest Daycare & 10 & ~ & ~ & $\checkmark$ \\  
        The Longest Daycare & 28 & ~ & ~ & $\checkmark$ \\  
        South Gondar Zone & 1 & ~ & $\checkmark$ & $\checkmark$ \\  
        South Gondar Zone & 4 & ~ & $\checkmark$ & ~ \\  
        South Gondar Zone & 16 & ~ & $\checkmark$ & ~ \\  
        South Gondar Zone & 3 & ~ & ~ & $\checkmark$ \\  
        Milton Friedman ... & 1 & ~ & ~ & $\checkmark$ \\  
        Milton Friedman ... & 8 & ~ & ~ & $\checkmark$ \\  
        Fedor Ozep & 2 & ~ & ~ & $\checkmark$ \\  
        TY.O & 1 & ~ & ~ & $\checkmark$ \\  
        TY.O & 3 & ~ & ~ & $\checkmark$ \\  
        TY.O & 10 & ~ & ~ & $\checkmark$ \\  
        TY.O & 13 & ~ & ~ & $\checkmark$ \\  
        TY.O & 14 & ~ & ~ & $\checkmark$ \\  
        TY.O & 20 & ~ & ~ & $\checkmark$ \\  
        TY.O & 29 & ~ & ~ & $\checkmark$ \\  
        Front of Islamic ... & 1 & ~ & ~ & $\checkmark$ \\ 
        Front of Islamic ... & 2 & ~ & ~ & $\checkmark$ \\  
        Front of Islamic ... & 3 & ~ & ~ & $\checkmark$ \\  
        Front of Islamic ... & 4 & ~ & ~ & $\checkmark$ \\  
        Front of Islamic ... & 5 & ~ & ~ & $\checkmark$ \\  
        Rufus Carter & 7 & ~ & $\checkmark$ & ~ \\  
        Rufus Carter & 8 & ~ & $\checkmark$ & ~ \\  
        Smoke Break & 1 & ~ & ~ & $\checkmark$ \\  
        Smoke Break & 2 & ~ & ~ & $\checkmark$ \\  
        Smoke Break & 3 & ~ & ~ & $\checkmark$ \\  
        Bambi II & 6 & ~ & $\checkmark$ & ~ \\  
        Bambi II & 8 & ~ & $\checkmark$ & ~ \\  
        Bambi II & 13 & ~ & $\checkmark$ & ~ \\  
        Bambi II & 15 & ~ & $\checkmark$ & ~ \\  
        Assassin's Creed Unity & 1 & ~ & $\checkmark$ & ~ \\  
        Assassin's Creed Unity & 16 & ~ & $\checkmark$ & ~ \\  
        Assassin's Creed Unity & 13 & ~ & $\checkmark$ & ~ \\  
        Assassin's Creed Unity & 3 & ~ & $\checkmark$ & ~ \\  
        Mehmet Çetingöz & 1 & ~ & $\checkmark$ & ~ \\  
        Mehmet Çetingöz & 2 & ~ & $\checkmark$ & ~ \\  
        Mehmet Çetingöz & 9 & ~ & $\checkmark$ & ~ \\  
        Mehmet Çetingöz & 12 & ~ & $\checkmark$ & ~ \\  
        Mehmet Çetingöz & 10 & ~ & $\checkmark$ & ~ \\  
        Baltimore and ... & ~ & ~ & ~ & $\checkmark$ \\  
        Baltimore and ... & ~ & ~ & ~ & $\checkmark$ \\  
        Dante Alighieri Society & 6 & ~ & ~ & $\checkmark$ \\  
        Osaka Bay & 25 & ~ & ~ & $\checkmark$ \\  
        Osaka Bay & 36 & ~ & ~ & $\checkmark$ \\  
        Osaka Bay & 38 & ~ & ~ & $\checkmark$ \\  
        Liang Congjie & 8 & ~ & $\checkmark$ & ~ \\  
        University (album) & 18 & ~ & ~ & $\checkmark$ \\  
        University (album) & 19 & ~ & ~ & $\checkmark$ \\  
        University (album) & 24 & ~ & ~ & $\checkmark$ \\  
        University (album) & 25 & ~ & ~ & $\checkmark$ \\  
        Lappeenranta & 2 & ~ & ~ & $\checkmark$ \\  
        Lappeenranta & 12 & ~ & ~ & $\checkmark$ \\  
        Lappeenranta & 13 & ~ & ~ & $\checkmark$ \\  
        Lappeenranta & 18 & ~ & ~ & $\checkmark$ \\  
        Ali Abdullah Ahmed & 4 & ~ & ~ & $\checkmark$ \\  
        Ali Abdullah Ahmed & 8 & ~ & ~ & $\checkmark$ \\  
        Ali Abdullah Ahmed & 9 & ~ & ~ & $\checkmark$ \\  
        Joseph R. Anderson & 9 & ~ & $\checkmark$ & ~ \\  
        Ramblin' on My Mind & 1 & ~ & ~ & $\checkmark$ \\  
        Ramblin' on My Mind & 2 & ~ & $\checkmark$ & ~ \\  
        \bottomrule
    \end{tabular}}
    \caption{The category of each error in the original DocRED.}
    \label{error4}
\end{table}

\begin{table}
    \centering
    \scalebox{0.65}{
    \begin{tabular}{l|c|c|c|c}
    \toprule
        Document Title & Rel. & Err. 1 & Err. 2 & Err. 3 \\ \hline 
        Christopher Franke & 3 & ~ & ~ & $\checkmark$ \\  
        Christopher Franke & 4 & ~ & $\checkmark$ & ~ \\  
        Christopher Franke & 5 & ~ & $\checkmark$ & ~ \\  
        Statue of Jan ... & 3 & ~ & ~ & $\checkmark$ \\  
        Statue of Jan ... & 4 & ~ & ~ & $\checkmark$ \\  
        Robert Taylor & 1 & ~ & ~ & $\checkmark$ \\  
        Robert Taylor & 2 & ~ & ~ & $\checkmark$ \\  
        Robert Taylor & 3 & ~ & ~ & $\checkmark$ \\  
        Robert Taylor & 4 & ~ & ~ & $\checkmark$ \\  
        Sycamore Canyon & 1 & ~ & $\checkmark$ & ~ \\  
        Amos Hochstein & 9 & ~ & $\checkmark$ & ~ \\  
        Paul Pfeifer & 3 & ~ & $\checkmark$ & ~ \\  
        Mega Man Zero & 8 & ~ & $\checkmark$ & ~ \\  
        Soldier (song) & 1 & ~ & ~ & $\checkmark$ \\  
        Soldier (song) & 3 & ~ & ~ & $\checkmark$ \\  
        Gloria Estefan ... & 4 & ~ & $\checkmark$ & ~ \\  
        Anthony G. Brown & 3 & ~ & ~ & $\checkmark$ \\  
        Harbour Esplanade & 3 & ~ & $\checkmark$ & ~ \\  
        Harbour Esplanade & 5 & ~ & $\checkmark$ & ~ \\  
        Harbour Esplanade & 6 & ~ & $\checkmark$ & ~ \\  
        Henri de Buade & 3 & $\checkmark$ & ~ & ~ \\  
        The Reverent ... & 5 & $\checkmark$ & ~ & ~ \\  
        This Little ... & 6 & ~ & ~ & $\checkmark$ \\  
        This Little ... & 9 & ~ & ~ & $\checkmark$ \\  
        This Little ... & 13 & ~ & ~ & $\checkmark$ \\  
        This Little ... & 15 & ~ & ~ & $\checkmark$ \\  
        This Little ... & 19 & ~ & ~ & $\checkmark$ \\  
        This Little ... & 20 & ~ & ~ & $\checkmark$ \\  
        Ali Akbar Moradi & 1 & ~ & ~ & $\checkmark$ \\  
        Ali Akbar Moradi & 2 & ~ & ~ & $\checkmark$ \\  
        Ali Akbar Moradi & 3 & ~ & ~ & $\checkmark$ \\  
        I Knew You ... & 2 & ~ & ~ & $\checkmark$ \\  
        I Knew You ... & 4 & ~ & ~ & $\checkmark$ \\  
        I Knew You ... & 5 & ~ & ~ & $\checkmark$ \\  
        I Knew You ... & 6 & ~ & ~ & $\checkmark$ \\  
        I Knew You ... & 7 & ~ & ~ & $\checkmark$ \\  
        I Knew You ... & 8 & ~ & ~ & $\checkmark$ \\  
        I Knew You ... & 9 & ~ & ~ & $\checkmark$ \\  
        Mohammed A. W. & 6 & ~ & $\checkmark$ & ~ \\  
        Mohammed A. W. & 8 & ~ & $\checkmark$ & ~ \\  
        Mohammed A. W. & 10 & ~ & $\checkmark$ & ~ \\  
        Elbląg County & 5 & ~ & ~ & $\checkmark$ \\  
        The Crazy World ... & 1 & ~ & $\checkmark$ & ~ \\  
        The Crazy World ... & 6 & ~ & $\checkmark$ & ~ \\  
        The Crazy World ... & 7 & ~ & $\checkmark$ & ~ \\  
        The Crazy World ... & 8 & ~ & $\checkmark$ & ~ \\  
        The Crazy World ... & 9 & ~ & $\checkmark$ & ~ \\  
        Flag of Prussia & 1 & ~ & ~ & $\checkmark$ \\  
        Flag of Prussia & 3 & $\checkmark$ & ~ & ~ \\  
        Flag of Prussia & 7 & ~ & ~ & $\checkmark$ \\  
        Flag of Prussia & 11 & ~ & $\checkmark$ & ~ \\  
        John Alexander Boyd & 11 & ~ & ~ & $\checkmark$ \\  
        John Alexander Boyd & 12 & ~ & ~ & $\checkmark$ \\  
        Municipal elections ... & 5 & ~ & ~ & $\checkmark$ \\  
        Municipal elections ... & 7 & ~ & ~ & $\checkmark$ \\  
        House of Angels & 7 & ~ & ~ & $\checkmark$ \\  
        William James Wallace & 7 & ~ & ~ & $\checkmark$ \\  
        William James Wallace & 8 & ~ & $\checkmark$ & ~ \\  
        William James Wallace & 10 & ~ & $\checkmark$ & ~ \\  
        William James Wallace & 11 & ~ & $\checkmark$ & ~ \\  
        Black Mirror (song) & 7 & ~ & ~ & $\checkmark$ \\  
        Michael Claassens & 5 & ~ & ~ & $\checkmark$ \\  
        Michael Claassens & 12 & ~ & ~ & $\checkmark$ \\  
        Lark Force & 13 & ~ & ~ & $\checkmark$ \\  
        Washington Place & 9 & ~ & ~ & $\checkmark$ \\  
        Battle of Chiari & 2 & ~ & ~ & $\checkmark$ \\  
        Battle of Chiari & 6 & ~ & ~ & $\checkmark$ \\  
        Woodlawn, Baltimore ... & 18 & ~ & ~ & $\checkmark$ \\  
        Wagner–Rogers Bill & 1 & ~ & ~ & $\checkmark$ \\  
        \bottomrule
    \end{tabular}}
    \caption{The category of each error in the original DocRED.}
    \label{error5}
\end{table}

\end{document}